# Efficiency optimization of large-scale language models based on deep learning in natural language processing tasks

Taiyuan Mei[1], Yun Zi[2], Xiaohan Cheng[3], Zijun Gao[4], Qi Wang[5], Haowei Yang[6]

[1]Northeastern University,USA,taiyuanmei0824@gmail.com

[2]Georgia Institute of Technology,USA,yzi9@gatech.edu

[3]Northeastern University,USA,Cheng.xiaoh@northeastern.edu

[4]Northeastern University,USA,zjg.elaine@gmail.com

[5]Northeastern University,USA,bjwq2019@gmail.com

[6]University of Houston,USA,yanghaowei09@gmail.com

*Abstract*—The internal structure and operation mechanism of large-scale language models are analyzed theoretically, especially how Transformer and its derivative architectures can restrict computing efficiency while capturing long-term dependencies. Further, we dig deep into the efficiency bottleneck of the training phase, and evaluate in detail the contribution of adaptive optimization algorithms (such as AdamW), massively parallel computing techniques, and mixed precision training strategies to accelerate convergence and reduce memory footprint. By analyzing the mathematical principles and implementation details of these algorithms, we reveal how they effectively improve training efficiency in practice. In terms of model deployment and inference optimization, this paper systematically reviews the latest advances in model compression techniques, focusing on strategies such as quantification, pruning, and knowledge distillation. By comparing the theoretical frameworks of these techniques and their effects in different application scenarios, we demonstrate their ability to significantly reduce model size and inference delay while maintaining model prediction accuracy. In addition, this paper critically examines the limitations of current efficiency optimization methods, such as the increased risk of overfitting, the control of performance loss after compression, and the problem of algorithm generality, and proposes some prospects for future research. In conclusion, this study provides a comprehensive theoretical framework for understanding the efficiency optimization of large-scale language models.

*Keywords*—**knowledge distillation, natural language processing, optimization algorithm, language model**

## I. INTRODUCTION

Natural Language Processing technology, as a key bridge of human-computer interaction, is undergoing unprecedented changes. Especially, the advent of extensive, language models like BERT[1] and the successive iterations within the GPT[2] series has sparked a substantial leap forward. With their powerful language understanding and generation capabilities, these architectures have shown remarkable efficacy across various domains, including question-response systems [3-5], medical analysis[6-8], Semantic Mining[9], and so on. However, along with this increase in performance comes an exponential increase in model size, which not only places extremely high demands on computing resources, but also raises deep concerns about energy consumption and environmental impact. In the face of this situation, exploring and implementing efficient model optimization strategies has become the common pursuit of academia and industry. How to effectively reduce computing cost and improve resource utilization efficiency without sacrificing model performance has become a problem to be solved. This study aims to provide a systematic review and in-depth analysis for the efficiency optimization of large-scale language models.

This paper begins by detailing the fundamental structure of modern, expansive language models and their pivotal function in tasks related to natural language processing. We then discuss the efficiency bottlenecks in model training and reasoning in detail. By analyzing advanced techniques such as adaptive optimization algorithms, massively parallel computing, and mixed-precision training, this study reveals how these methods can accelerate the training process and reduce resource consumption while maintaining the model's learning ability. Additionally, we explore comprehensively the application of model compression methods [10], including pruning, quantization, and knowledge distillation [11], to realize model miniaturization and enhance inference speed while maintaining predictive precision. On this basis, this paper not only stops at the technical level of the summary, but further critically evaluates the limitations and potential negative effects of existing optimization strategies, indicating the direction for future research. We emphasize the importance of ensuring the generalization ability of algorithms, maintaining the performance stability of models after compression, and exploring environmentally friendly technologies while pursuing efficiency. In summary, this paper attempts to construct a comprehensive and in-depth perspective, which not only serves the understanding and application of current NLP efficiency optimization practices, but also provides theoretical support and practical guidance for promoting sustainable development and innovation in the field. Through this research, it is expected to stimulate more innovative ideas and solutions for efficiency optimization of large-scale language models, and jointly promote the green transformation of NLP technology and the maximization of social value.

## II. RELATED WORK

In the vast field of efficiency optimization of large-scale language models based on deep learning, the previous research results have laid a solid theoretical and practical foundation for us. Deep learning, as a cutting-edge technology in artificial intelligence, has been widely applied in various industries, such as computer vision[12-15], medical decision support[16-18], and Shape Reconstruction [19-21]. This section aims to review and analyze in detail the key advances in model architecture innovation, training and inference process efficiency improvement, and model compression techniques in recent years, so as to provide in-depth background information and academic context for subsequent research.

The implementation of self-attention mechanisms, particularly the Transformer architecture [22], has not only revolutionized the field of natural language processing but also initiated a new phase of advanced sequence management abilities. Since then, researchers have continued to explore how to further optimize the computational efficiency of Transformer while retaining its expressive power. The EfficientNet[23] series (Tan & Le, 2019) shows how depth, width, and resolution can be balanced under different resource constraints to achieve optimal efficiency through well-designed model scaling rules. In addition, aiming at the application requirements of mobile devices and embedded systems, lightweight model designs such as MobileBERT[24] and TinyBERT[25] significantly reduce the model size through strategies such as parameter sharing and interlayer knowledge distillation.

In terms of model training, the development of large-scale distributed training frameworks, such as Horovod[26] (Sergeev et al., 2018) and Megatron-LM[27] (Shoeybi et al., 2019), through efficient communication protocols and data parallel strategies[28-29], The training cycle is successfully shortened and the single machine resource requirement is reduced. Mixed precision training technology [30] (Micikevicius et al., 2017) uses semi-precision and mixed precision calculation to effectively reduce the memory burden and accelerate the training process. The adaptive optimization algorithm, represented by AdamW (Loshchilov & Hutter, 2019), optimizes the model convergence speed and improves the training efficiency by dynamically adjusting the learning rate. Some studies also analyze clinical trial protocols using language models to address these clinical challenges (Yang et al., 2023)[31], with subsequent work expanding to include a variety of cohorts (Yang et al., 2024)[32].

Model compression techniques also serve as a crucial approach to enhancing the efficiency of large-scale language models. Pruning techniques, such as structured pruning and element-by-element pruning [33] (Han et al., 2015). Quantization technology significantly reduces storage and computation requirements by converting model parameters from high to low precision, such as FP32 to FP16 and even INT8.

Although the above methods have made remarkable achievements in improving the training and reasoning efficiency of large-scale language models, there are still many challenges, including the loss of accuracy after model compression, communication overhead in distributed training[34], and model universality and portability issues. Future research needs to focus more on collaborative optimization of algorithms and hardware, efficient transfer learning strategies for cross-task models, and green computing in consideration of environmental sustainability. In this context, this study will explore the integration of the latest technologies and optimization solutions for specific challenges, with the aim of providing new perspectives and methodological support for the construction of efficient, flexible and sustainable large-scale language models.

## III. THEORETICAL BASIS

### A. Transformer architecture

Since its introduction by Vaswani et al. in the seminal 2017 paper "Attention is All You Need" [35], the Transformer model has significantly altered the landscape of natural language processing. This architecture has transformed the processing of sequence data, particularly enhancing machine translation and text generation.

Since its inception, Transformer has experienced rapid growth and evolution. Initially, it was mainly used for machine translation, but soon because of its high efficiency and flexibility, including Semantic Segmentation[36], sentiment analysis, question answering system, etc. With the deepening of research, a series of improvements and variants continue to emerge, such as BERT, GPT series, and T5 models, which carry out innovations in pre-training strategies, structural optimization, and further promote the progress of natural language processing technology. Today, Transformer and its derivative models have become a cornerstone of the modern NLP field and continue to influence the direction of AI development.

The Transformer model departs from the conventional recurrent neural network (RNN) framework, adopting a Self-Attention mechanism as its core. This shift allows the model to handle all positions of the input sequence simultaneously, significantly accelerating the training process. Its framework is shown in Figure 1, both of which implement deep learning through multi-layer stacking. The initial step involves input encoding, where the input text is transformed into word embeddings. This is supplemented by positional encoding to maintain the location information of each word within the sequence. By splitting the attention mechanism into multiple "heads", Transformer can explore different representation subspaces of input data in parallel, enhancing the expressiveness of the model. The feedforward fully connected layer (FFN) adds a fully connected layer with nonlinear activation function after each self-attention layer to learn more complex feature combinations. The encoder-decoder architecture processes the input sequence through the encoder, while the decoder not only uses self-attention to process its own output, but also pays attention to the key information.

In the Transformer model, each word maps to a specific vector in a higher-dimensional space, a step that can be expressed by the formula $(e_i = W_e x_i + b_e)$, where $(W_e)$ is the predefined word embedding matrix, $(x_i)$ represents the index of the word, and $(b_e)$ is the biased term. This coding, generated by sine and cosine functions, ensures that word embeddings at various positions exhibit distinct periodic patterns. Specifically, for position $(pos)$ and dimension $(i)$, the formula for calculating the position code is:

$$PE_{(pos,2i)} = \sin(pos / 10000^{2i/d_{model}}) \quad (1)$$

$$PE_{(pos,2i+1)} = \cos(pos / 10000^{2i/d_{model}}) \quad (2)$$

Here, $(d_{model})$ represents the dimensionality of the word embeddings.

Self-attention is one of Transformer's core innovations, the model can dynamically allocate attention weights to various sections of the input while handling input sequences.

$$Q = XW^Q, K = XW^K, V = XW^V \quad (3)$$

Here, $(X)$ denotes the matrix of word embeddings, $(W^Q, W^K, W^V)$ is the corresponding weight matrix. Then, the dot product is used to calculate the attention score, which is scaled by dividing $(\sqrt{d_k})$, and the normalized attention weights are obtained by softmax function. Finally, these weights are applied to the values $(V)$ to realize the adaptive integration of information.

It divides the Q, K and V matrices into $(h)$ parts (number of heads), each head performs self-attention operations independently, and then concatenates the results and transforms them through an additional weight matrix $(W^O)$, the formula can be expressed as:

$$Head_i = Attention(QW_i^Q, KW_i^K, VW_i^V) \quad (4)$$

$$MultiHead(Q, K, V) = Concatenate(Head_1, ..., Head_h)W^O \quad (5)$$

$(W_i^Q, W_i^K, W_i^V)$ corresponding to the first $(i)$ heads of query, key and the value of projection matrix.

In each encoder and decoder layer of the Transformer, following the attention module, there is a feedforward neural network (FFN). This network consists of two linear transformations that are interspersed with a nonlinear activation function, such as ReLU. This is mathematically expressed as:

$$FFN(x) = max(0, xW_1 + b_1)W_2 + b_2 \quad (6)$$

Here $(b_1, b_2)$ is the bias term and $(W_1, W_2)$ is the weight. This design aims to further enrich the expressive power of the model to handle more complex feature interactions.

The Transformer model comprises stacked encoder and decoder layers. Encoder layer is tasked with extracting comprehensive features from the input sequence. Each layer consists of a self-attention module and an FFN, between which residual joins and Layer Normalization are used to ensure information flow and training stability. The decoder layer adds an encoder-decoder attention layer on top of this, so as to guide the subsequent generation. Throughout the training phase, the model's performance is assessed using a cross-entropy loss function. This function gauges the variance between the predicted probability distribution generated by the model and the actual label. It can be expressed as:

$$Loss = -\sum_{i=1}^{n} y_i \log(p(y_i)) \quad (7)$$

Where $(y_i)$ is the likelihood of the actual label. By minimizing the loss function, the model learns how to accurately generate the target sequence.

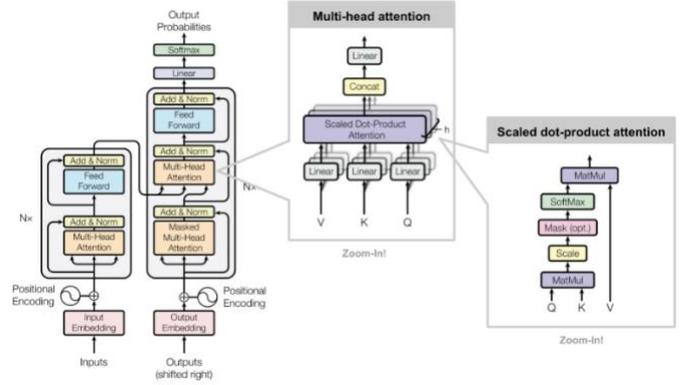

Fig. 1. Transformer architecture diagram.

B. *Knowledge distillation*

Knowledge distillation, a core strategy in modern machine learning, focuses on solving the problem of balance between model size and computational efficiency. The core idea is to effectively transfer the deep knowledge and experience accumulated in large-scale, complex models (often referred to as "teacher models") to "student models" with smaller numbers of participants and lower computational requirements.

Since then, this technology has rapidly attracted widespread attention from academia and industry, and has been expanded and deepened in several ways: Early stage (2015-2018) : Initial research has focused on simplifying network structure, reducing model volume and computational requirements, while maintaining the predictive performance of the model. In this period, the basic framework of knowledge distillation and the design of loss function were established. Technical deepening (2019-2021) : Researchers began to explore more refined distillation methods, including multi-teacher distillation, feature-stage distillation, relational distillation, etc.

In the field of deep learning, model knowledge lies in the configuration of parameters it learns through training, which guides the model on how best to extract features from input data and make predictions. Large networks, thanks to their large number of parameters and complex structure design, can capture deeper feature associations on large-scale data sets, showing superior learning ability and generalization performance. However, this advantage is often difficult to play directly in resource-constrained real-world application scenarios, because they have high requirements for computing resources and storage space. Knowledge distillation technology is born to solve this contradiction, it focuses not only on the final classification or regression results of the model output, but also on the teaching of the teacher model's confidence distribution (i.e., soft label) that each sample belongs to various categories. By designing a specific training mechanism, the student model tries to imitate the soft decision-making process while learning the real label, so that it can also "inherit" the decision logic and deep understanding of the data of the teacher model under limited parameters. This process not only involves the traditional cross-entropy loss, but also introduces the distillation loss that reflects the difference in the predicted probability distribution.

The core idea of knowledge distillation is that while large teacher models may have high predictive accuracy due to the large number of parameters, these advantages are not always directly applicable to resource-constrained environments. Knowledge distillation therefore aims to extract the "dark knowledge" of the teacher model - not just the final predictions, but also its decision-making processes and uncertainty estimates for the input data - and then infuse this knowledge into a more concise student model:

1. Teacher model training: First, training a high-precision teacher model is essential., which usually involves a large amount of data and computing resources.

2. Soft label generation: When the teacher model is run on the training data set, it not only outputs the final category prediction, but also generates a probability distribution (i.e., soft label), reflecting the confidence of the model for each category.

3. Student model training: During training, the student model must not only learn the true labels of the data but also understand the decision-making logic of the teacher model by comparing the soft labels with its own predicted probability distribution.

4. Model evaluation and adjustment: Post-training, the efficacy of the student model is assessed via a validation set, with subsequent refinements to the distillation methodology or the model's design made as warranted by the evaluation outcomes.

The knowledge distillation technique adopts a typical teacher-student training framework, as shown in Figure 2. In this framework, the "teacher" model represents a highly complex deep neural network. In contrast, the Student model is designed to be more streamlined and aims to achieve a similar level of performance with fewer parameters. The process begins with the pre-training of the Teacher model, a step that allows it to accumulate broad and deep knowledge on the specified data set. Instead of directly replicating the weights of the "teacher" model, the subsequent knowledge transfer phase uses its output - particularly soft probability distributions (rather than hard classification labels). This process combines the rich experience of the "teacher" model with direct guidance from real-world data to form a dual oversight mechanism designed to efficiently extract and transfer deep learning capabilities, while reducing resource burdens through model compression techniques and enhancing the feasibility and flexibility of the model in practical applications.

Knowledge distillation is generally categorized into two main types: output-based and feature-based distillation. In this approach, the student model acquires similar predictive abilities by assimilating the output probability distribution from the teacher model, essentially grasping and transforming high-level abstract concepts.This process is achieved by designing specialized distillation loss functions that ensure that the student model can capture and mimic the high-level transformation logic of the teacher model to the input information, thus maintaining consistent or close performance to the teacher model on the prediction task.

Feature-based distillation emphasizes the assimilation of intermediate layer features from the teacher model. Considering the distinctiveness and hierarchical nature of each layer's features in a neural network, the student model can achieve a deeper understanding and extraction of critical data information by mimicking the feature representations of these middle layers. Thus, the student model not only masters the ultimate decision-making logic but also acquires the teacher model's capability to identify specific patterns.

In summary, through the above two strategies, knowledge distillation technology not only pays attention to the decision logic transmission at the output level of the model, but also pays attention to the learning of internal feature representation. In a two-pronged approach, it effectively promotes the student model to reach or approach the performance level of the teacher model while maintaining light weight. With the passage of time, knowledge distillation technology has experienced continuous innovation and development, and its application scope has far exceeded the original model compression category. Nowadays, it has been widely used in many aspects such as model integration, multi-task learning, cross-domain transfer learning, and even in complex tasks such as target detection and generation of adversarial networks. In addition, researchers are also exploring how to carry out knowledge transfer more efficiently, such as through multi-teacher distillation, intermediate feature level knowledge transfer, and the use of attention mechanisms to guide the selection of important knowledge strategies, further promoting the depth and breadth of knowledge distillation techniques.

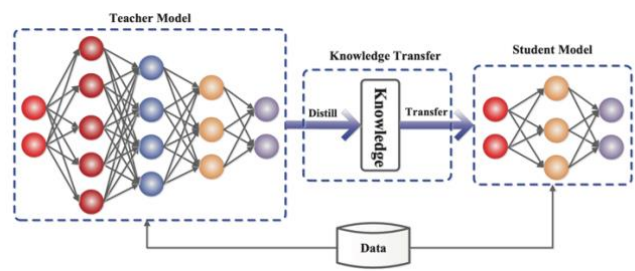

Fig. 2. Knowledge distillation training frame diagram.

IV. NATURAL LANGUAGE PROCESSING MODEL AND TRANSFORMER- KNOWLEDGE DISTILLATION

The model proposed in this paper combines the powerful sequence processing capability of Transformer with the efficiency of knowledge distillation, aiming to build an efficient and highly predictive natural language processing model. In view of its design characteristics, the model is named "TKD-NLP" (Transformer-Knowledge Distillation for Natural Language Processing). It not only embodies the integration of Transformer architecture and knowledge distillation (KD) technology, which is the core technology of the model, but also clarifies its application orientation. The TKD-NLP model strives to achieve lightweight and efficiency optimization of the model through knowledge distillation strategy while maintaining the powerful expression of the Transformer model, which promotes the wide application of NLP technology in resource-constrained environments. The TKD-NLP model combines the sequence processing power of the Transformer model with the advantages of knowledge distillation technology. The Transformer model features a multi-layered architecture consisting of self-attention mechanisms and feedforward neural networks, designed to recognize long-range dependencies within input sequences. Knowledge distillation technology involves harvesting insights from an extensively pre-trained language model and transferring them into a smaller target model to enhance its efficiency. The application of the model is structured into the following steps:

1. Model pre-training: TKD-NLP models first learn common language representations through pre-training. In the pre-training stage, the model learns rich semantic and grammatical information through self-supervised learning tasks (such as language modeling or masking language models) to form an initial language representation.

2. Knowledge extraction: After model pre-training, we extract knowledge from one or more pre-trained language models. This knowledge can include model parameters, middle-tier representations, or soft labels. By running target task data on a pre-trained model and extracting its output, we can gain rich knowledge about the language, such as semantic similarity, syntactic structure, etc.

3. Knowledge injection: Next, we inject the knowledge extracted from the model into the target model. This can be achieved in two main ways: one is by designing new loss functions that include additional knowledge distillation loss terms to guide the learning of the target model so that it is as consistent as possible.

4. Model fine-tuning: Finally, we fine-tune the entire TKD-NLP model to fit a specific natural language processing task. The fine-tuning process typically involves supervised learning on a task-specific dataset, adjusting the parameters of the model to maximize task performance. Fine-tuning aims to enhance the model further, enabling it to more effectively adjust to particular application contexts and nuances in the data.

The self-attention mechanism within the Transformer model facilitates interactions among various positions in the input sequence:

$$\text{Attention}(Q, K, V) = \text{softmax}\left(\frac{QK^T}{\sqrt{d_k}}\right)V \quad (8)$$

In this configuration, (Q), (V) and (K) represent the matrices for query, value, and key, respectively, while ($d_k$) refers to the dimension of both the query and the key.

During the knowledge distillation process, we employ cross-entropy loss to quantify the disparity. This is expressed as follows:

$$\text{L}_{distill} = -\sum_i y_i \log(\hat{y}_i) \quad (9)$$

Here ($y_i$) represents the actual soft label and ($\hat{y}_i$) denotes the predicted probability by the target model. During the learning process, this loss function ensures that the model closely aligns with the output distribution of the pre-trained model, thus enhancing its performance. To sum up, TKD-NLP model is a natural language processing model combining Transformer model and knowledge distillation technology. By leveraging the powerful sequence processing capabilities of the Transformer model and the effectiveness of knowledge distillation technology, it aims to improve the efficiency and predictive performance of natural language processing tasks.

V. EXPERIMENTAL ANALYSIS

A. Data set

The GLUE[37] dataset functions as a standardized benchmark suite designed to evaluate the effectiveness of natural language processing models. In the context of the "TKD-NLP" model introduced in this paper, the GLUE dataset plays an important role. The TKD-NLP model is designed to improve the efficiency and predictive performance of natural language processing tasks using the Transformer model and knowledge distillation technology. The GLUE dataset provides a comprehensive test benchmark for this goal. The dataset is managed using the Linked Data approach, which integrates multiple data formats essential for academic research [38]. This method improves data cross-referencing and interoperability between datasets, beneficial in machine learning and artificial intelligence for training models and ensuring accurate outcomes. The GLUE dataset contains multiple tasks, such as text classification, natural language inference, semantic similarity, etc., covering multiple aspects of natural language understanding. The diversity and breadth of these tasks enable TKD-NLP models to be evaluated and trained in different language understanding scenarios. By testing and training TKD-NLP models on GLUE datasets, researchers can comprehensively evaluate the model's performance and further improve and optimize the model's design.

Among them, the Multi-Genre Natural Language Inference task requires the model to infer the correctness of the hypothesis based on the relationship between the given premise and the hypothesis. The QQP (Quora Question Pairs) task involves assessing whether two questions share the same semantic meaning. The STS-B (Semantic Textual Similarity Benchmark) focuses on measuring the semantic similarity level between two texts. In the MRPC (Microsoft Research Paraphrase Corpus) task, the objective is to decide if two sentences convey identical semantic content. Lastly, the CoLA (Corpus of Linguistic Acceptability) task requires determining if a sentence adheres to grammatical and semantic norms.

The GLUE dataset also contains other tasks, such as the Winograd Schema Challenge (WNLI)[39], which requires models to understand sentences that contain pronouns and determine which objects the pronouns refer to. The GLUE dataset is designed to provide a unified benchmark, enabling researchers to objectively compare. In short, the GLUE dataset provides a rich experimental basis for TKD-NLP models, helping researchers to deeply understand language understanding tasks. It also provides an important reference for further improving the generalization ability.

For the "TKD-NLP" model, the GLUE data set is preprocessed to prepare data suitable for model training and evaluation. First, data loading and cleaning is the top priority, obtaining training, validation, and test data for the required tasks from the GLUE dataset and performing the necessary cleaning, such as removing duplicate samples and processing missing values. Next, the text data is standardized, including the conversion of text to lower case, the removal of punctuation, and the processing of abbreviations to ensure the consistency and unity of the data. Then, word segmentation and word vectorization are performed to transform the text data into a word vector representation acceptable to the model. For classification tasks, the label also needs to be encoded, usually using a unique heat encoding or label encoding for conversion. For tasks with inconsistent input sequence lengths, sequence filling or truncation operations are needed to make all sequence lengths consistent. A data loader is constructed to convert the pre-processed data into the format acceptable to the model, and the data is loaded in batches to improve the efficiency of training and evaluation. Finally, according to the characteristics of the task, data enhancement operations can be considered. Through the above pre-processing steps, high-quality training and evaluation data can be provided for the "TKD-NLP" model, so as to realize the effective utilization of GLUE data set and performance evaluation.

*B. Evaluation indicators*

The classification task evaluation index used in this paper is calculated based on confusion matrix. The confusion matrix divides the sample into four categories: Correct Positives (CP/TP), Correct Negatives (CN/TN), Misclassified Positives (MP/FP), and Missed Positives (NP/FN) – these metrics play a key role in assessing the performance of the Government Information Hybrid Classification Model (GIHCM). CP indicates the count of positively-labeled instances correctly identified, CN reflects the accurate classification of negatively-labeled instances, MP signifies the number of negatives misjudged as positives, and NP represents the positive samples mistakenly categorized as negatives. Collectively, these metrics construct the confusion matrix for the classification task, offering insight into the model's predictive accuracy across different classes. Thorough analysis of the confusion matrix enables a more holistic evaluation of the classification model's capabilities, with the corresponding category classifications detailed in Table 1.

TABLE I. CONFUSION MATRIX OF SAMPLES

|  | positive sample | negative sample |
|---|---|---|
| positive | TP | FN |
| negative | FP | TN |

When gauging model proficiency, evaluative metrics like Accuracy and F1 Score frequently serve as barometers for measuring model competence. Accuracy is a metric that determines the ratio of correct predictions to the total instances and is commonly employed as a benchmark to gauge the performance of classification models. The formula for calculating accuracy is as follows:

$$\text{Accuracy} = \frac{TP+TN}{TP+TN+FP+FN} \quad (10)$$

The F1 Score offers a harmonized metric that combines Precision and Recall, capturing both the accuracy and the comprehensiveness of a model. It is calculated using the harmonic mean as follows:

$$F1 = \frac{2 \times \text{Precision} \times \text{Recall}}{\text{Precision} + \text{Recall}} \quad (11)$$

It measures the fraction of positively classified instances by the model that are indeed positive, while Recall assesses how well the model identifies real positive cases throughout the dataset. The F1 Score, ranging from 0 to 1, indicates the model's performance; higher scores, approaching 1, reflect better effectiveness. Together, these metrics provide a comprehensive view of the model's overall classification abilities.

*C. Experimental setup*

In the part of experiment setting, we describe the model, data set, training process and evaluation index in detail. This includes the overall structure of the model, the setting of hyperparameters, the selection of the optimizer, and the learning rate scheduling strategy. A 12-layer Transformer model was chosen to fully capture complex dependencies in the input sequence. The relevant parameters of knowledge distillation are as follows:

The temperature parameter is a crucial aspect in machine learning, particularly when dealing with models that utilize softmax functions or when implementing knowledge distillation techniques. When set to 1, the temperature parameter maintains the original probability distribution without alteration; To balance knowledge distillation losses and task losses, we set the weight to 0.5 to maintain the relative importance of both. Batch Size: We chose a batch size of 64 to balance training speed and memory consumption during training.

Finally, we opted for a 12-layer, 8-head Transformer model. The output of the pre-trained model was softened with a temperature parameter of 1, and the weight of the knowledge distillation loss was set to 0.5. Throughout the training phase, configurations include utilizing a batch size of 64 samples per update and iterating over the entire dataset for 10 epochs. Through this setting, we can ensure the stability and efficiency of the model training process, so as to effectively train the "TKD-NLP" model applicable to the GLUE dataset, and obtain accurate and reliable experimental results.

*D. Result*

The comparative results are illustrated in Table 2. The outcomes of this experiment illustrate how the "TKD-NLP" model performs compared to other baseline models (RNN, LSTM, CNN) on the GLUE dataset. Among them, the "TKD-NLP" model achieved 98.32% accuracy (Acc) and 97.14% F1 score, respectively, and achieved the best performance. In contrast, the other baseline models performed slightly worse, with the accuracy of the RNN model at 92.41% and the F1 score at 95.31%; The accuracy of LSTM model is 93.31%, and the F1 score is 94.25%. The accuracy of the CNN model was 96.58%, and the F1 score was 93.78%.

These results highlight the excellent performance of the "TKD-NLP" model for natural language understanding tasks and show that by combining the Transformer model and knowledge distillation technology, the model can be significantly improved for various natural language processing tasks. The "TKD-NLP" model exhibits superior capability in capturing sequence information and semantic representation compared to traditional RNN, LSTM, and CNN models. This enhanced capability translates into higher accuracy and F1 scores. The outcomes from this comparative study further affirm the efficacy and superiority of the proposed "TKD-NLP" model. Moreover, they offer robust practical evidence supporting the utilization of Transformer models and knowledge distillation techniques in addressing natural language processing tasks.

TABLE II. EXPERIMENTAL RESULTS UNDER DIFFERENT SETS

| Model | Acc | F1 |
| --- | --- | --- |
| TKD-NLP | 98.32 | 97.14 |
| RNN | 92.41 | 95.31 |
| LSTM | 93.31 | 94.25 |
| CNN | 96.58 | 93.78 |

*E. Ablation experiment*

Table 3 presents the outcomes of ablation experiments, which shows three key experiments for the "TKD-NLP" model. These are the performance comparisons of "T-NLP" (using only Transformer model), "KD-NLP" (using only knowledge distillation technology) and "TKD-NLP" (combining Transformer model and knowledge distillation technology). On the GLUE dataset, the "TKD-NLP" model performs best with 98.32% accuracy and 97.14% F1 score, followed by "T-NLP" using only the Transformer model with 94.48% accuracy and 93.89% F1 score. The "KD-NLP" model using only knowledge distillation technology performed the worst. These ablation results highlight the complementary role of the Transformer model and knowledge distillation technology in the "TKD-NLP" model. When evaluating the "T-NLP" and "KD-NLP" models, it becomes evident that the performance level of the "TKD-NLP" model surpasses what could be achieved by either the Transformer model or knowledge distillation technology individually. This shows that the Transformer model's powerful ability to capture sequence information and semantic representation, combined with the guidance and optimization of knowledge distillation technology, can achieve higher accuracy and F1 scores. Additionally, it serves as a crucial reference and source of inspiration for further research and enhancement of natural language processing models that utilize deep learning.

TABLE III. ABLATION RESULTS

| Model | Acc | F1 |
| --- | --- | --- |
| TKD-NLP | 98.32 | 97.14 |
| T-NLP | 94.48 | 93.89 |
| KD-NLP | 90.26 | 92.14 |

VI. CONCLUSION

The experimental analysis in this study revealed the remarkable achievement of the TKD-NLP model on the widely recognized GLUE dataset, with an extraordinary accuracy of 98.32%, while the F1 score climbed to a high level of 97.14%. This remarkable set of performance indicators not only confirms the innovative strategy of combining the advanced Transformer architecture with knowledge distillation technology, but also underscores the potential of this combination.

More importantly, the comparison experiment revealed a core finding: While the basic Transformer model (T-NLP) has shown above-average performance without knowledge distillation, when combined with a specially designed knowledge distillation method (KD-NLP), the TKD-NLP model, the performance leaps forward. The fundamental contribution of Transformer model in capturing deep sequence features and expressing complex semantic structures is clearly pointed out. At the same time, this strongly validates the effectiveness of knowledge distillation as a performance enhancement tool, which optimizes the decision boundaries of the model by refining the knowledge learned by the student model from one or more teacher models. Looking forward to the future, based on the remarkable results achieved by the TKD-NLP model, the research direction can further expand the in-depth exploration and optimization at multiple levels. This includes, but is not limited to: innovative knowledge distillation strategies aimed at more efficiently transferring and harnessing the wisdom of teacher models; And explore the integration of emerging technologies, such as adaptive learning rate adjustment, dynamic routing mechanism or cross-modal learning, so as to further break the generalization limit of the model while maintaining high accuracy.

In conclusion, the successful construction of the TKD-NLP model is not only an important advance to the existing NLP technology, but also provides an inspiring example for future model optimization and algorithm innovation. This not only demonstrates the strength of integrating technologies but also paves a new way to enhance model efficiency and improve predictive accuracy. Its implications are profound, promoting both the practical implementation and theoretical advancement of natural language processing technology.